\pdfoutput=1

\documentclass[11pt]{article}

\usepackage[]{ACL2023}

\usepackage{multirow}
\usepackage{multicol}
\usepackage{tcolorbox}
\usepackage{enumitem}
\usepackage{xcolor,colortbl}
\usepackage{booktabs}
\usepackage{xspace}
\newcommand{\MEDICint}{{MEDIC}\xspace}
\newcommand{\MEDICext}{{MEDIC}$^\complement$\xspace}

\newcommand{\SNOMEDint}{{SNOMED}\xspace}
\newcommand{\SNOMEDext}{{SNOMED}$^\complement$\xspace}

\newcommand{\TAint}{{T038}\xspace}
\newcommand{\TAext}{{T038}$^\complement$\xspace}

\newcommand{\TBint}{{T058}\xspace}
\newcommand{\TBext}{{T058}$^\complement$\xspace}

\usepackage{refstyle}
\usepackage{times}
\usepackage{latexsym}
\usepackage{amssymb}
\usepackage{amsmath}
\usepackage{cleveref}
\crefformat{section}{\S#2#1#3}
\crefformat{subsection}{\S#2#1#3}
\crefformat{subsubsection}{\S#2#1#3}
\crefrangeformat{section}{\S\S#3#1#4 to~#5#2#6}
\crefmultiformat{section}{\S\S#2#1#3}{ and~#2#1#3}{, #2#1#3}{ and~#2#1#3}
\crefmultiformat{subsection}{\S\S#2#1#3}{ and~#2#1#3}{, #2#1#3}{ and~#2#1#3}

\Crefformat{figure}{#2Fig.~#1#3}
\Crefmultiformat{figure}{Figs.~#2#1#3}{ and~#2#1#3}{, #2#1#3}{ and~#2#1#3}
\Crefformat{table}{#2Tab.~#1#3}
\Crefmultiformat{table}{Tabs.~#2#1#3}{ and~#2#1#3}{, #2#1#3}{ and~#2#1#3}
\Crefformat{appendix}{Appx.~\S#2#1#3}
\crefmultiformat{appendix}{Appx.~\S#2#1#3}{ and~#2#1#3}{, #2#1#3}{ and~#2#1#3}
\crefformat{algorithm}{Alg.~#2#1#3}
\usepackage[T1]{fontenc}

\usepackage[utf8]{inputenc}

\usepackage{microtype}

\usepackage{inconsolata}

\title{Exploring Partial Knowledge Base Inference in Biomedical Entity Linking}

\author{Hongyi Yuan\textsuperscript{\protect\includegraphics[width=0.3cm]{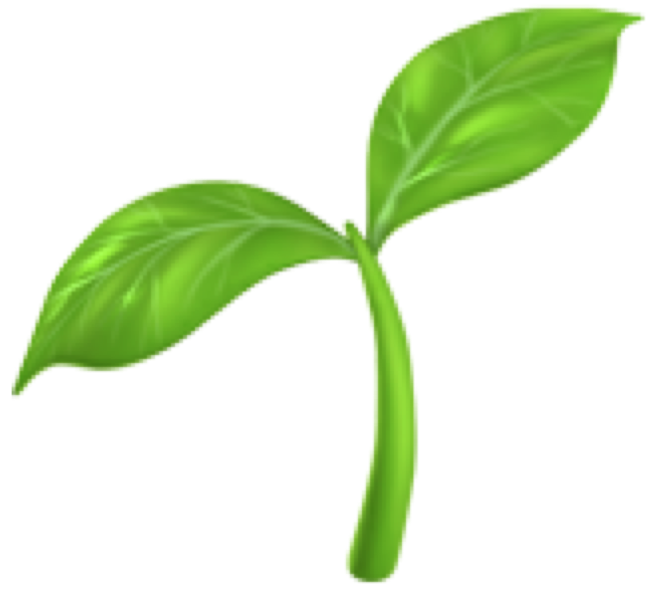}}  \\
  Tsinghua University \\
  \texttt{yuanhy20@mails.tsinghua.edu.cn} \\\And
  Keming Lu\textsuperscript{\protect\includegraphics[width=0.3cm]{emoji1.png}} \\
  University of Southern California \\
  \texttt{keminglu@usc.edu} \\
  \AND
  Zheng Yuan\textsuperscript{\protect\includegraphics[width=0.3cm]{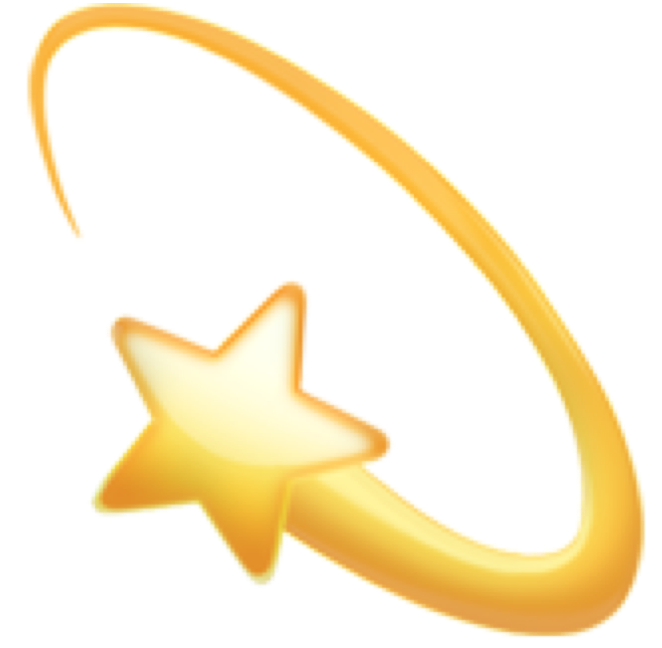}}\thanks{\protect\includegraphics[width=0.3cm]{emoji2.png} Corresponding Author. \protect\includegraphics[width=0.3cm]{emoji1.png} Contributed equally. Ordering is determined by dice rolling.} \\
  Alibaba Group \\
  \texttt{yuanzheng.yuanzhen@alibaba-inc.com} \\}
\renewcommand\footnotemark{}

\begin{document}
\maketitle

\begin{abstract}
Biomedical entity linking (EL) consists of named entity recognition (NER) and named entity disambiguation (NED).
EL models are trained on corpora labeled by a predefined KB.
However, it is a common scenario that only entities within a subset of the KB are precious to stakeholders.
We name this scenario partial knowledge base inference: training an EL model with one KB and inferring on the part of it without further training.
In this work, we give a detailed definition and evaluation procedures for this practically valuable but significantly understudied scenario and evaluate methods from three representative EL paradigms.
We construct partial KB inference benchmarks and witness a catastrophic degradation in EL performance due to dramatically precision drop.
Our findings reveal these EL paradigms can not correctly handle unlinkable mentions (NIL), so they are not robust to partial KB inference. 
We also propose two simple-and-effective redemption methods to combat the NIL issue with little computational overhead. Codes are released at \url{https://github.com/Yuanhy1997/PartialKB-EL}.
\end{abstract}

\section{Introduction}
Biomedical entity linking (EL) aims to identify entity mentions from biomedical free texts and link them to the pre-defined knowledge base (KB, e.g. UMLS~\cite{bodenreider2004unified}), which is an essential step for various tasks in biomedical language understanding including relation extraction~\cite{li2016biocreative,lin2020high,hiai2021relation,yu2022bios} and question answering~\cite{qareview}.

\begin{figure}
    \centering
    \small
     \includegraphics[width=0.48\textwidth]{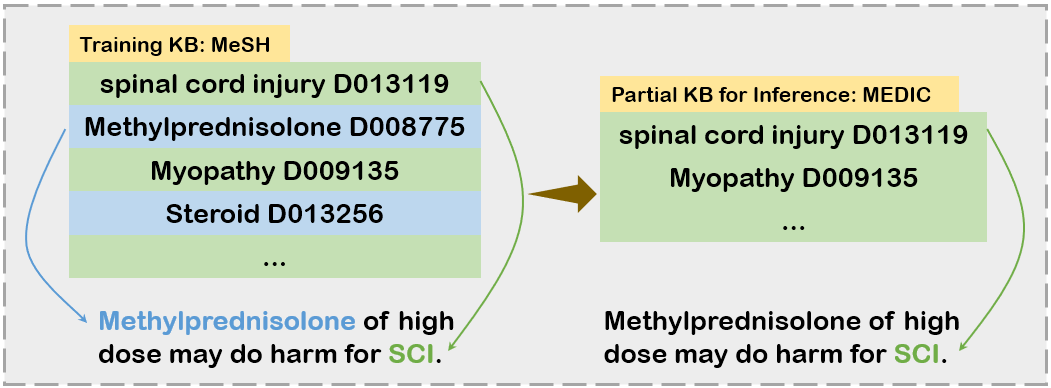}
    \caption{Visual illustration of partial KB inference scenario. Partial KB inference from training KB MeSH (left) to a partial KB MEDIC (right). \textit{Methylprednisolone} is not extracted since it is not in MEDIC.}
    \label{fig:task_overview}
    \vspace{-1em}
\end{figure}

EL naturally contains two subtasks: named entity recognition (NER) and named entity disambiguation (NED).
NER is designed for mention detection, while NED aims to find the best match entities from KB.
One direct way for EL is executing NER and NED sequentially~\cite{liu2020self,zhang2021knowledge,yuan2022coder}.
Neural NER and NED models are usually trained by corpora labeled with a KB.
However, potential users of biomedical EL, including doctors, patients, and developers of knowledge graphs (KGs) may only be interested in entities inside a subset of KB such as SNOMED-CT \cite{donnelly2006snomed}, one semantic type of entities in UMLS, or KB customized by medical institutions.
Besides, doctors from different medical institutions have different terminology sets. 
Some hospitals are using ICD-10, while some hospitals are still using ICD-9 and even custom terminology set.
Patients are only interested in specific diseases, symptoms, and drugs.
As for developers of KGs, they may need to build a KG for special diseases like diabetes \cite{chang2021diakg} and COVID-19 \cite{reese2021kg}, or particular relation types like drug-drug interaction \cite{lin2020kgnn}. 
All scenarios above need to infer EL using a partial KB.
Off-the-shelf models trained on a comprehensive KB will extract mentions linked to entities outside the users' KB.
Although retraining models based on users' KBs can obtain satisfactory performances, it is not feasible under most scenarios because users can have significantly different KBs and may have difficulties with computational resources in finetuning large-scale models.
Therefore, we propose a scenario focusing on inference on the partial KB.
We name this scenario \textbf{partial knowledge base inference}: Train an EL model with one KB and infer on partial of this KB without further training.
\Cref{fig:task_overview} provides a case of this scenario.
This scenario is widely faced in the medical industry but remains understudied.



This work reviews and evaluates current state-of-the-art EL methods under the partial KB inference scenario.
To be specific, we evaluate three paradigms:
(1) \textit{NER-NED} \cite{yuan-etal-2021-improving,coder},
(2) \textit{NED-NER} \cite{zhang2022entqa},
(3) \textit{simultaneous generation} \cite{genre}.
The first two paradigms are pipeline methods, whose difference is the order of NER and NED.
The last paradigm is an end-to-end method that generates mention and corresponding concepts by language models.
We construct partial KB inference datasets based on two widely used biomedical EL datasets: BC5CDR \cite{li2016biocreative} and MedMentions \cite{medmentions}.
Our experimental findings reveal the different implicit mechanisms and performance bottlenecks within each paradigm which shows partial KB inference is challenging.

We also propose two redemption methods based on our findings, \textbf{post-pruning} and \textbf{thresholding}, to help models improve partial KB inference performance effortlessly.
Post-pruning infers with a large KB and removes entities in the large KB but not in partial KB.
Post-pruning is effective but memory-unfriendly for storing embeddings of entities in the large KB.
Thresholding removes entities with scores below a threshold.
These two redemption methods are all designed to reduce the impact of NIL entities and boost EL performances.
To our best knowledge, this is the first work that researches partial KB inference in biomedical EL.
Our main contributions are the following:

\begin{itemize}[leftmargin=1em]
    \setlength\itemsep{0em}
    \item We extensively investigate partial KB inference in biomedical EL. We give a detailed definition, evaluation procedures, and open-source curated datasets.
    
    \item Experiment results show that the NED-NER paradigm behaves more robust towards partial KB inference, while the other paradigms suffer from sharp degradation caused by NIL.
    
    \item We propose two redemption techniques to address the NIL issue with little computational overhead for better partial KB inference.
    
\end{itemize}

\section{Related Work}

\paragraph{NER and NED} In biomedical and general domains, NER and NED are two extensively studied sub-fields in NLP. As mentioned, EL can be decomposed and approached by NER and NED.
NER is often considered a sequential labeling task \cite{lample2016neural}.
Neural encoders like LSTM \cite{gridach2017character,10.1093/bioinformatics/btx228,cho2019biomedical} or pretrained language models \cite{weber2021hunflair} encode input text and assign BIO/BIOES tags to each word. 
Many biomedical pretrained language models are proposed to enhance NER performances \cite{beltagy-etal-2019-scibert,peng2019transfer,lee2020biobert,gu2021domain,yuan-etal-2021-improving}.
Concerning NED, most methods embed mentions and concepts into a common dense space by language models and disambiguate mentions by nearest neighbor search \cite{dualencoder,biocom,rescnn}. 
\citet{clustering} and \citet{arboel} first rerank the disambiguation target to boost performance. To overcome the limitation of labeled NED corpus, \citet{sapbert,coder,yuan-etal-2022-generative} leverage synonyms from huge biomedical KB for zero-shot NED.
\citet{dataintegration,krissbert} use weakly supervised data generated from Wikipedia and PubMed for data augmentation. 
NER and NED are both essential components of EL.
In this work, we further explore partial KB inference by analyzing performance in these two steps and reveal how the design and order of NER and NED infer EL performance in partial KB inference.

\paragraph{Entity Linking} Although EL can be handled by a direct pipeline of NER and NED, there is limited research focusing on the task as a whole in biomedical. 
As EL may enjoy the mutual benefits from supervision of both subtasks, \citet{Zhao_Liu_Zhao_Wang_2019} deal with biomedical EL in a multi-task setting of NER and NED.
MedLinker \cite{medlinker} and \citet{ujiie-etal-2021-end} approach biomedical EL by sequentially dealing with NER and NED using a shared language model and they devise a dictionary-matching mechanism to deal with concepts absent from the training annotations. 

In the general domain, GENRE \cite{genre,mgenre} is proposed and formulated EL as a seq2seq task. They detect and disambiguate mentions with constrained language generation in an end-to-end fashion. We categorize GENRE as \textit{simultaneous-generate} EL.
EntQA \cite{zhang2022entqa} provides a novel framework by first finding probable concepts in texts and then treating each extracted concept as queries to detect corresponding mentions in a question-answering fashion which is categorized as \textit{NED-NER} in our framework. 
\textit{Simultaneous-generate} and \textit{NED-NER} fashion are not widely examined in biomedical EL, and they interest us to examine their performances for biomedical EL and partial KB inferences.

\paragraph{Partial KB inference in EL}
In the biomedical domain, there is no prior work considering this setting to the best of our knowledge.
NILINKER \cite{nilinker} is the most related work which focuses on linking NIL entities out of the training KB, while ours aim to infer EL on part of the training KB and discard NIL entities.




\begin{figure*}[t]
    \centering
    \includegraphics[width=0.9\linewidth]{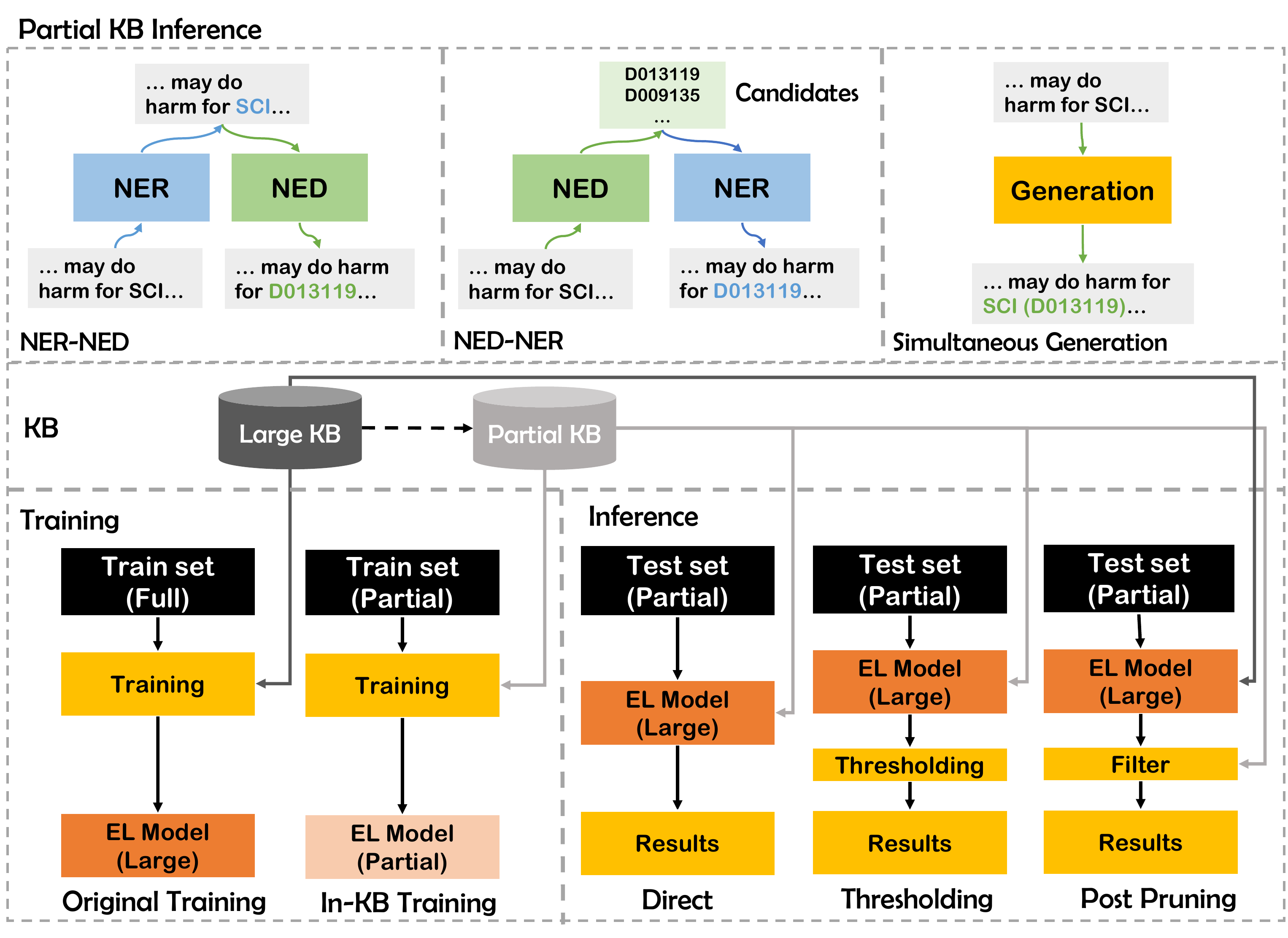}
    \caption{Overview of three different entity linking paradigms and settings of partial KB inference.
    The top sub-graph demonstrates three EL paradigms we investigated in this work (\Cref{sec:methods}).
    The middle sub-graph shows the relation of the large training KB and partial KB in inference (\Cref{sec:definition}).
    The bottom sub-graph shows two EL models obtained from full and partial training and three partial KB inference settings.
    The direct partial KB inference is the naive setting described in \Cref{sec:methods}.
    Thresholding and post pruning are two simple redemption methods we propose and describe in \Cref{sec:post-processing}.}
    \label{fig:method_overview}
    \vspace{-1em}
\end{figure*}

\section{Problem Definition}\label{sec:definition}

\paragraph{Entity Linking} Let $\mathcal{E}$ denote a target KB comprises of a set of biomedical concepts.
Given a text $\boldsymbol{s}$ with length $n$, an EL model aims to find the mentions $m$ and corresponding concepts $e\in \mathcal{E}$.
Concretely, the model can be regarded as a mapping $f:\boldsymbol{s}\to\mathcal{P}_\mathcal{E}$, where $\mathcal{P}_\mathcal{E} = \{(i, j, e)|0\leq i\leq j\leq n, e\in\mathcal{E}\}$ denotes the possible target mention-concept pairs, and $i,j$ mark the start and end positions of the mention spans in $\boldsymbol{s}$.


\paragraph{Partial KB inference} In the conventional EL scenario, the target KB is the same in training and inference. 
In this paper, we consider a partial KB inference scenario containing two different KBs, $\mathcal{E}_1$ and $\mathcal{E}_2$, and assume $\mathcal{E}_1\supsetneq\mathcal{E}_2$.
The larger KB $\mathcal{E}_1$ corresponds to the training KB while the smaller KB $\mathcal{E}_2$ corresponds to the \textit{partial inference KB}.
Models are required to map a text $\boldsymbol{s}$ to a different label set $\mathcal{P}_{\mathcal{E}_2}$ during inference, rather than $\mathcal{P}_{\mathcal{E}_1}$ during training, and we have $\mathcal{P}_{\mathcal{E}_1}\supsetneq\mathcal{P}_{\mathcal{E}_2}$. 
There exists a label distribution shift in this scenario.
We investigate whether current entity linking models are robust for partial KB inference and how models perform under the shifted distribution of targets. 


\section{Experiments}

In this section, we introduce our experimental setup,
which includes implementation details of EL methods we investigated (\Cref{sec:methods}) and datasets we create for investigating partial KB inference (\Cref{sec:datasets}).

\begin{table*}[th]
\small 
\centering
\setlength{\tabcolsep}{1.5mm}{
\begin{tabular}{p{1mm}ll|ccc|ccc|ccc}
\toprule
& \multicolumn{2}{c}{Target KB.}& \multicolumn{3}{c}{EntQA} & \multicolumn{3}{c}{GENRE}& \multicolumn{3}{c}{KeBioLM+CODER} \\
& Train KB & Eval KB &Precision & Recall & F1&Precision & Recall & F1&Precision & Recall & F1 \\ 
\hline
\cellcolor{blue!10}& UMLS & UMLS  & 45.99&23.68&31.27&42.44&43.69&43.05&33.58&34.94&34.25\\
\cellcolor{blue!10}&&\SNOMEDint  & 46.04&27.01&34.05&34.40&49.40&40.56&28.19&48.28&35.59\\
\cellcolor{blue!10}&&\SNOMEDext  & 
36.75 & 23.12 & 28.38 & 19.82 & 39.28 & 26.35 & 14.18 & 37.54 & 20.59 \\
\cellcolor{blue!10}&&\TAint  & 41.52&31.56&35.86&17.26&49.53&25.60&9.78&50.28&16.37\\
\cellcolor{blue!10}&&\TAext  & 
43.43 & 23.24 & 30.28 & 34.97 & 42.45 & 38.35 & 26.52 & 34.59 & 30.02 \\
\cellcolor{blue!10}& &\TBint  & 30.01&25.56&27.61&7.69&36.06&12.68&4.76&41.51&8.54\\
\cellcolor{blue!10}\multirow{-7}{*}{\rotatebox[origin=c]{90}{\textit{MedMentions}}} &&\TBext  & 
 46.02 & 24.34 & 31.84 & 40.45 & 44.76 & 42.50 & 31.95 & 37.74 & 34.61 \\
\hline
\multicolumn{3}{c}{Avg. Drop}\vline & 5.36 & -2.13 & -0.7 & 16.68 & 0.11 & 12.04 & 14.35 & -6.71 & 9.96 \\
\hline
\cellcolor{blue!10}&MeSH&MeSH & 83.59 & 66.48 & 74.06 &70.92&68.71&69.80&72.21&74.84&73.5 \\
\cellcolor{blue!10}&&\MEDICint  & 81.92&70.45&75.75&31.53&68.19&43.12&29.24&68.38&40.96\\
\cellcolor{blue!10}\multirow{-3}{*}{\rotatebox[origin=c]{90}{\textit{BC5.}}}&&\MEDICext & 
87.10 & 66.92 & 75.69 & 37.55 & 65.33 & 47.69 & 42.57 & 80.67 & 55.73 \\
\hline
\multicolumn{3}{c}{Avg. Drop}\vline & -0.92 & -2.21 & -1.66 & 36.38 & 1.95 & 24.40 & 36.31 & 0.32 & 25.16 \\
\bottomrule
\end{tabular}}
\caption{Results for entitly linking in parital KB inference. The first section shows results on MedMentions with UMLS as training KB. The last section shows results on BC5CDR with MeSH as training KB. Eval KB represents different partial KBs for inference. 
The average drops are averaged among metrics between full evaluation (first row in each section) and partial KB evaluation (other rows).}
\label{tab:main_end-to-end_result}
\small 
\vspace{-1em}
\end{table*}

\subsection{Direct Partial KB Inference}\label{sec:methods}
There are three widely-used paradigms for entity linking: (1) \textit{NER-NED}; (2) \textit{NED-NER}; (3) \textit{Simultaneous Generation}.
We introduce representative methods for each paradigm and how methods are accommodated to partial KB inference with minimal change. 
\textbf{To be noticed, these paradigms are not aware of the KB $\mathcal{E}_1$ during partial KB inference.}
The top subgraph in \Cref{fig:method_overview} depicts the overview of the three paradigms.
We also describe how directly applying these methods to partial KB inference, which corresponds to the Direct inference method in \Cref{fig:method_overview}.
Hyper-parameters for experiments are reported in \Cref{app:train_hyper}.

\subsubsection{NER-NED}\label{sec:NER-NED}

A straightforward solution for entity linking is a two-phase paradigm that first detects the entity mentions by NER models and then disambiguates the mentions to concepts in KBs by NED models, shown in the left top subgraph of \Cref{fig:method_overview}.
We finetune a pre-trained biomedical language model for token classification as the NER model in this paradigm.
Specifically, we use KeBioLM \cite{yuan-etal-2021-improving} as our language model backbone.
We use CODER \cite{yuan2022coder} as our NED model which is a self-supervised biomedical entity normalizer pre-trained on UMLS synonyms with contrastive learning.
CODER disambiguates mentions by generating embedding from each concept synonym and recognized mentions into dense vectors and then finding the nearest concept neighbors of each mention vector by maximum inner product search (MIPS).

In partial KB inference, although the NER model is not aware of the changes in KB, the NED model only needs to search for the nearest concept within a partial KB.
Smaller inference KB is challenging for the NED model. 
For a mention $m$ and its corresponding concept $e\in\mathcal{E}_1$, if $e\notin\mathcal{E}_2$, the NED model will return an incorrect or less accurate concept from $\mathcal{E}_2$. 
Since the users are only interested in concepts within $\mathcal{E}_2$, these kinds of mention $m$ should be linked as unlinkable entities (NIL).

\subsubsection{NED-NER}\label{sec:NED-NER}

NED-NER methods are also formatted as a  two-phase pipeline, which is shown in the middle top subgraph of \Cref{fig:method_overview}.
This paradigm first retrieves the concepts mentioned in the text, then identifies mentions based on retrieved concepts. 
This paradigm is proposed along with the method EntQA \cite{zhang2022entqa}. 
In the concept retrieval phase of EntQA, a retriever finds top-K related concepts for texts by embedding both into a common dense space using a bi-encoder, then searches nearest neighbors for texts by MIPS within the partial KB $\mathcal{E}_2$.
This phase retrieves concepts from raw texts directly and we view it as the NED phase.
Following its original setting, We initialize the retriever from BLINK \cite{wu2019scalable} checkpoints and further fine-tune the bi-encoder on our datasets with its contrastive loss functions.
In the following phase, a reader is trained to identify mentions in a question-answering fashion where mentions and concepts correspond to answers and queries respectively. 
This phase is viewed as NER.
In partial KB inference, only concepts from the partial KB  will be encoded into dense vectors for MIPS. 



\subsubsection{Simultaneous Generation}\label{sec:generation}

In the generative paradigm for entity linking, NER and NED are achieved simultaneously, which is shown in the right top subgraph of \Cref{fig:method_overview}.
Entity linking is modeled as a sequence-to-sequence (seq2seq) task where models insert special tokens and concept names into texts with a constrained decoding technique via a Trie.
We follow the detailed model design in GENRE.
Given a input text $\boldsymbol{s}$, the target sequence is built as: $s^{\text{tar}} = \{\ldots,M^B,x_i,\ldots x_j,M^E,E^B,e,E^E,\ldots\}$,
where $x_i,\ldots x_j$ are the mention tokens in $\boldsymbol{s}$, $e$ is a token sequence of the concept name, and $M^B,M^E,E^B,E^E$ are special tokens marking the beginning and ending of mentions and concepts. 
The model is trained in seq2seq fashion by maximum log-likelihood with respect to each token. 
During inference, a token prefix trie is built to constrain model only output concepts within the given KB. 
For partial KB inference, only concept names from the partial KB are added to build the prefix Trie in GENRE. 
This will ensure all entity linking results will only be referred to the partial KB.

\subsection{Datasets}\label{sec:datasets}

We conduct experiments on two widely-used biomedical EL datasets and select several partial KBs used as inference.
Selection biases of partial KBs may be introduced into our setting because different partial KBs may result in different target distributions of mention-concept annotations, as this may lead to different difficulties in EL due to different KB sizes, the semantics of entities, and entity occurrence frequencies in the training set.
To eliminate this effect as much as possible, we not only evaluate on partial KBs mentioned above but also their complement KBs to the training KBs.
We add $\complement$ to indicate the complements. 
The detailed statistics of datasets are listed in \Cref{tab:stats} of \Cref{app:data_stat}.

\textbf{BC5CDR}~\cite{li2016biocreative} is a dataset that annotates 1,500 PubMed abstracts with 4,409 chemicals, 5818 disease entities, and 3,116 chemical-disease interactions. 
All annotated mentions are linked to concepts in the target knowledge base MeSH.
We use MeSH as the training KB and we consider a smaller KB \textit{MEDIC} \cite{davis2012medic} as the partial KB for inference.
MEDIC is a manually curated KB composed of 9,700 selected disease concepts mainly from MeSH.

\textbf{MedMentions} \cite{medmentions} is a large-scale biomedical entity linking datasets curated from annotated PubMed abstracts.
We use the \textit{st21pv} subset which comprises 4,392 PubMed abstracts, and over 350,000 annotated mentions linked to concepts of 21 selected semantic types in UMLS \cite{bodenreider2004unified}.
We use UMLS as the training KB and we select three representative partial KBs which are concepts from semantic types \textit{T038 (Biologic Function)} and \textit{T058 (Health Care Activity)} in UMLS and \textit{SNOMED}.


\section{Results}

In this section, we present the main results of partial KB inference (\Cref{sec:main-result})
Then, we provided two redemption methods for enhancing model performance in partial KB inference (\Cref{sec:post-processing}).
In the end, we discuss the factors related to difficulties hindering partial KB inference performance (\Cref{sec:analysis}).

\begin{table*}[t]
\small 
\centering
\setlength{\tabcolsep}{1.5mm}{
\begin{tabular}{p{2mm}ll|ccc|ccc|ccc}
\toprule
& \multicolumn{2}{c}{Target KB.}& \multicolumn{3}{c}{EntQA} & \multicolumn{3}{c}{GENRE}& \multicolumn{3}{c}{KeBioLM+CODER} \\
&Train KB & Eval KB &Precision & Recall & F1&Precision & Recall & F1&Precision & Recall & F1 \\ 
 \hline
\cellcolor{blue!10}&UMLS &UMLS  & 82.72 &51.81&63.72&64.27&66.17&65.21&69.08&71.88&70.45\\
\cellcolor{blue!10}&&\SNOMEDint  & 82.09 & 51.83 & 63.54 & 45.78&65.74&53.97&43.22&74.04&54.58\\
\cellcolor{blue!10}&&\SNOMEDext  & 80.57 &53.82 &64.53 & 30.34 & 60.13 & 40.33 & 26.94 & 71.32 & 39.11 \\
\cellcolor{blue!10}&&\TAint  & 82.43&52.66&64.27&22.34&64.10&33.13&14.85&76.36&24.86\\
\cellcolor{blue!10}&&\TAext  & 82.08&51.83&63.54& 53.44 & 64.86 & 58.60 & 55.26 & 72.07 & 62.56 \\
\cellcolor{blue!10}&&\TBint  & 78.92& 56.54 &65.88&11.37&53.31&18.75&7.76&67.68&13.93\\
\cellcolor{blue!10}\multirow{-7}{*}{\rotatebox[origin=c]{90}{\textit{Medmentions}}}&&\TBext  &82.91 &50.53&62.78& 59.90 & 66.30 & 62.94 & 62.35 & 73.65 & 67.53 \\
\hline
\multicolumn{3}{c}{Avg. Drop}\vline & 1.22 & -1.06 & -0.37 & 27.08 & 3.76 & 20.59 & 34.02 & -0.64 & 26.68 \\
\hline 
\cellcolor{blue!10}&MeSH&MeSH & 94.67 &82.56 &88.20 &87.59&84.86&86.20&86.47&91.05&88.70 \\
\cellcolor{blue!10}&&\MEDICint  & 92.31 &84.04 &87.99 & 37.85&81.84&51.76 &36.46&86.46&51.29\\
\cellcolor{blue!10}\multirow{-3}{*}{\rotatebox[origin=c]{90}{\textit{BC5.}}}&&\MEDICext & 96.37 & 82.93 & 89.14 & 49.07 & 85.38 & 62.32 & 50.13 & 94.99 & 65.63 \\
\hline
\multicolumn{3}{c}{Avg. Drop}\vline & 0.33 & -0.93 & -0.37 & 44.13 & 1.25 & 29.16 & 43.18 & 0.33 & 30.24 \\
\bottomrule
\end{tabular}}
\caption{Results for mention detection in partial KB inference.  Table arrangements are the same as \Cref{tab:main_end-to-end_result}.
}
\label{tab:mention_detect}
\small 
\vspace{-1em}
\end{table*}

\begin{table}[t]
\small 
\centering
\setlength{\tabcolsep}{0.5mm}{
\begin{tabular}{p{2mm}ll|cccc}
\toprule
&\multicolumn{2}{c}{Target KB.}& \multicolumn{2}{c}{EntQA} & \multicolumn{1}{c}{GENRE}& \multicolumn{1}{c}{Ke.+CO.} \\
&Train KB & Eval KB & R@100 & Acc. & Acc. & Acc. \\ 
 \hline
\cellcolor{blue!10}&UMLS &UMLS  & 57.26 & 75.38 &66.03&48.61\\
\cellcolor{blue!10}&&\SNOMEDint  & 65.86&74.81&75.14&65.22\\
\cellcolor{blue!10}&&\SNOMEDext  & 61.72& 68.67 &65.33 & 52.64\\
\cellcolor{blue!10}&&\TAint  & 75.10&65.34&77.26&65.86\\
\cellcolor{blue!10}&&\TAext  &58.54 & 66.89 &65.44 &47.99 \\
\cellcolor{blue!10}&&\TBint  & 74.28&57.92&67.63&61.34\\
\cellcolor{blue!10}\multirow{-7}{*}{\rotatebox[origin=c]{90}{\textit{MedMentions}}}&&\TBext  &58.76 & 68.52 &67.53 &51.24 \\
\hline
\multicolumn{3}{c}{Avg. Drop} & -8.45 & 8.35 & -3.69 & -8.77 \\
\hline 
\cellcolor{blue!10}&MeSH&MeSH & 80.34 & 92.72 & 80.97& 83.51 \\
\cellcolor{blue!10}&&\MEDICint  & 88.72 & 90.95 & 83.30 & 80.20\\
\cellcolor{blue!10}\multirow{-3}{*}{\rotatebox[origin=c]{90}{\textit{BC5.}}}&&\MEDICext  & 77.73 & 93.23 &76.52  &84.92 \\
\hline
\multicolumn{3}{c}{Avg. Drop} & -2.89 & 0.63 & 1.06 & 0.95\\
\bottomrule
\end{tabular}
}
\caption{Results for NED in partial KB inference. The disambiguation accuracies (Acc.) are calculated with respect to correctly detected mentions. For EntQA, we additionally add recall at the top 100 (R@100) to show its first-stage concept retrieval performance.}
\label{tab:main_dismbi}
\small 
\vspace{-1em}
\end{table}

\begin{table*}[h]
\small 
\centering
\setlength{\tabcolsep}{1.2mm}{
\begin{tabular}{p{2mm}l|cccc|ccccc}
\toprule
&& \multicolumn{4}{c}{\MEDICint}\vline & \multicolumn{4}{c}{\MEDICext} \\
&& EL-P/R&EL-F1&NER-F1&NED-Acc& EL-P/R&EL-F1&NER-F1&NED-Acc \\
\hline
\cellcolor{blue!10}& In-KB Train &81.27/71.34&\textbf{75.98}& \textbf{88.16} & \textbf{92.14 }&86.87/69.30& \textbf{77.10}& \textbf{90.08} & \textbf{94.44}\\
\cellcolor{blue!10}& Partial KB Inference &81.92/70.45& \underline{75.75} &\underline{87.99} & \underline{90.95} &87.10/66.92 & \underline{75.69}&\underline{89.14}& \underline{93.23}\\
\cellcolor{blue!10}\multirow{-3}{*}{\rotatebox[origin=c]{90}{EntQA}}& \ \ w/ Post-pruning&62.97/64.99&63.96 &84.10 &80.76 & 80.02/63.11&70.57& 86.42 & 77.84\\
\hline
\cellcolor{blue!10}& In-KB Train &65.65/68.38& \underline{66.99} &\underline{78.56}& {85.26}&69.96/62.02& \underline{65.75}&\underline{85.52}&76.89 \\
\cellcolor{blue!10}& Partial KB Inference &31.53/68.19&43.12 &51.76&83.30& 37.55/65.33&47.69 &62.32 & 76.52 \\
\cellcolor{blue!10}& \ \ w/ Thresholding & 76.32/59.25&66.71&72.43 &\textbf{92.11} &69.05/56.99& 62.45 &74.86& \textbf{83.41} \\
\cellcolor{blue!10}\multirow{-4}{*}{\rotatebox[origin=c]{90}{GENRE}}& \ \ w/ Post-pruning & 69.31/68.59&\textbf{68.95} &\textbf{79.92}&\underline{86.27}&69.46/66.29& \textbf{67.83}&\textbf{86.47}&\underline{78.45}  \\
\hline
\cellcolor{blue!10}& In-KB Train  &63.98/68.47& 66.15 &\textbf{82.94} &80.48&77.52/80.65& \underline{79.05}&\textbf{92.82}&85.18 \\
\cellcolor{blue!10}& Partial KB Inference &29.24/68.38 &40.96 &51.29&80.20& 42.57/80.67&55.73 &65.63 & 84.92\\
\cellcolor{blue!10}& \ \ w/ Thresholding &79.20/65.08& \textbf{71.45}&78.46& \textbf{91.07}&86.32/77.04& \textbf{81.41} &83.35&\textbf{97.68} \\
\cellcolor{blue!10}\multirow{-4}{*}{\rotatebox[origin=c]{90}{Ke.+CO.}}& \ \ w/ Post-pruning&69.03/65.27& \underline{67.10} &\underline{78.48}&\underline{85.49}&69.17/80.67& 74.48&\underline{87.27}&\underline{85.34} \\
\bottomrule
\end{tabular}
}
\caption{Results of partial KB inference, In-KB training and two redemption methods for three investigated models. The results are evaluated on partial KB \MEDICint and \MEDICext in BC5CDR.
The best performance for a model in each dataset is identified with \textbf{bold} and the second is \underline{underlined}.
}
\label{tab:main_transfer_enhance}
\small 
\vspace{-1em}
\end{table*}

\subsection{Main Results}\label{sec:main-result}



\paragraph{EL}
\Cref{tab:main_end-to-end_result} shows entity linking results on different partial KB settings.
First of all, we witness a significant and consistent performance drop in precision among all methods on MedMentions.
EntQA has the least precision drop (5.36\%) while GENRE and KEBioLM+CODER have a more obvious decrease, which is 16.68\% and 14.35\%, respectively. 
On the opposite, recalls in partial KBs remained the same even slightly increased.
KeBioLM+CODER shows the largest average recall increase (6.71\%), followed by EntQA (2.13\%), while the average recall of GENRE remains the same (only drops 0.11\%). 
Due to the stability in precision, the average change of F1 by EntQA even slightly increases (-0.7\%). 
However, the average F1 of GENRE and KeBioLM+CODER drops significantly on partial KBs, which are 12.04\% and 9.96\%.
The same pattern appears in BC5CDR.
EntQA shows extraordinary robustness in direct partial KB inference in contrast to the degradation of GENRE and KeBioLM. 
For individual partial KBs, a consistent pattern of precision and F1 drop is observed in GENRE and KeBioLM+CODER and EntQA is more robust compared to others. 
The F1 degradation led to by precision decrease reflects that the models detect redundant mentions that are out of the partial KBs.

\paragraph{NER}
\Cref{tab:mention_detect} shows the results for mention detection.
When inference on partial KBs, both GENRE and KeBioLM+CODER show drastically F1 score decrease on mention detection.
On MedMentions, average drops are 20.59\% and 26.68\% respectively for GENRE and KeBioLM+CODER.
On BC5CDR, the average drops are 29.16\% and 30.24\%.
The large fluctuation mainly comes from the sharp decreases of mention detection precision which are 27.08\%/34.02\% on MedMentions, and 44.13\%/43.18\% for GENRE and KEBIOLM+CODER respectively.
By comparison, the recall barely changes on partial KB inference. 
On the contrary, the fluctuations for EntQA are marginal, and across metrics and datasets, the largest drop is only 3.80\% for precision. 
EntQA shows rather robust performance on mention detection.
The trend is consistent across different subset KBs. 
Generally, GENRE and KeBioLM+CODER are sensitive to the changes to partial KBs.
These models detect mentions in $\mathcal{E}_1-\mathcal{E}_2$ during inference.
Therefore, these two frameworks present large precision degradation while recall barely fluctuates. 
EntQA detects mentions relying on retrieved concepts from the first phase.
It learns to restrict mentions according to concepts so it behaves robustly in partial KB inference. 
The results indicate a main defect for NER-NED and simultaneous generative paradigms is that the reliance between concepts and mentions is not well modeled, hence having poor NER performance in partial KB inference.

\paragraph{NED}
\Cref{tab:main_dismbi} shows the performance on the NED for correctly detected mentions.
Disambiguation accuracy shows little fluctuation for all methods while slightly increases on MedMentions. 
For example, the accuracy of KeBioLM+CODER increases from 48.61\% to 65.86\% when KB transfers from UMLS to T038 semantic type.
These results reveal that models learn the mapping between related mentions and concepts and are not biased by the out-of-KB annotations. 
The shrunk concept space of partial KBs makes the disambiguation task easier and leads to performance improvement.

\paragraph{Conclusion} We can conclude that
(1) NER-NED and Generative frameworks are not robust to direct partial KB inference, while the performance of NED-NER framework is more stable;
(2) degradation of entity linking performance is mainly a result of drastically degenerated mention detection performance on partial KBs and entity disambiguation abilities are stable; 
(3) EntQA potentially handled NILs via filtering out irrelevant entities before NER, while other methods suffer from low precision due to mislinking NILs to existing entities.

\subsection{Simple Redemptions}
\label{sec:post-processing}

In former subsections, we identify performance drops in partial KB inference mainly due to precision drops in the mention detection.
We introduce two simple-yet-effective methods to redeem the performance drops for partial KB inference: \textbf{Post-pruning} and \textbf{Thresholding}, which are shown in \Cref{fig:method_overview} and an example is provided in \Cref{app:example}.
Two methods are motivated by removing NIL mentions for improving mention detection performances.

\paragraph{Post-Pruning} asks the model to infer using  $\mathcal{E}_1$ and remove mention-entity pairs  out of the partial KBs $\mathcal{E}_1 - \mathcal{E}_2$. \textbf{This redemption method is naive but needs to know $\mathcal{E}_1$.}

\paragraph{Thresholding} uses $\mathcal{E}_2$ for inference. After obtaining mention-entity pairs, it will search a fixed threshold $\theta$ on the development set to maximize F1 and remove results with scores under the threshold.
\textbf{This method is not aware of $\mathcal{E}_1$.}

Specifically, for KeBioLM+CODER, we set a threshold on the cosine similarities of detected mentions and their most similar concepts:
\begin{equation*}
    {\rm score}=\max_{e\in\mathcal{E}_2}\cos(h_m, h_e),
\end{equation*}
where $m$ represents the mention extracted by NER and $h$ represents embeddings.

For EntQA, we obtain $K$ entities from the retriever and we compute the score for $k^{th}$ entity with starting and ending index $s,t$:
\begin{equation*}
    {\rm score}=P_{re}(e_k|e_{1:K})P_{st}(s|e_k,\boldsymbol{s})P_{ed}(t|e_k,\boldsymbol{s})
\end{equation*}
where $P_{re}$ computes the probability of $e_k$ among all entities and $P_{st}$ and $P_{ed}$ computes probabilities of $s$ and $t$ are the starting and ending of $e_k$.
The original implementation of EntQA integrated thresholding during inference, so the partial KB inference is equivalent to inference with thresholding.

For GENRE, we use the log-likelihoods for the generated mention span and concepts names in the output sequences $\boldsymbol{s}_m^e = \{M^B,x_i,\ldots x_j,M^E,E^B,e,E^E\}$ as scores: 
\begin{equation*}
    {\rm score}=\frac{1}{|\boldsymbol{s}_m^e|}\sum_{x\in \boldsymbol{s}_m^e}\log(P_{\text{ar}}(x))
\end{equation*}
where $P_{\text{ar}}$ represents the token's probability auto-regressively conditioned on its preceding tokens.


We compare two methods with direct partial KB inference.
We also include a setting where models are trained on the partial KB $\mathcal{E}_2$ for comparison.
We dub this `in-domain' setting as \textbf{In-KB train}.

\begin{figure*}[ht]
    \centering
    \resizebox{0.9\textwidth}{!}{
    \includegraphics{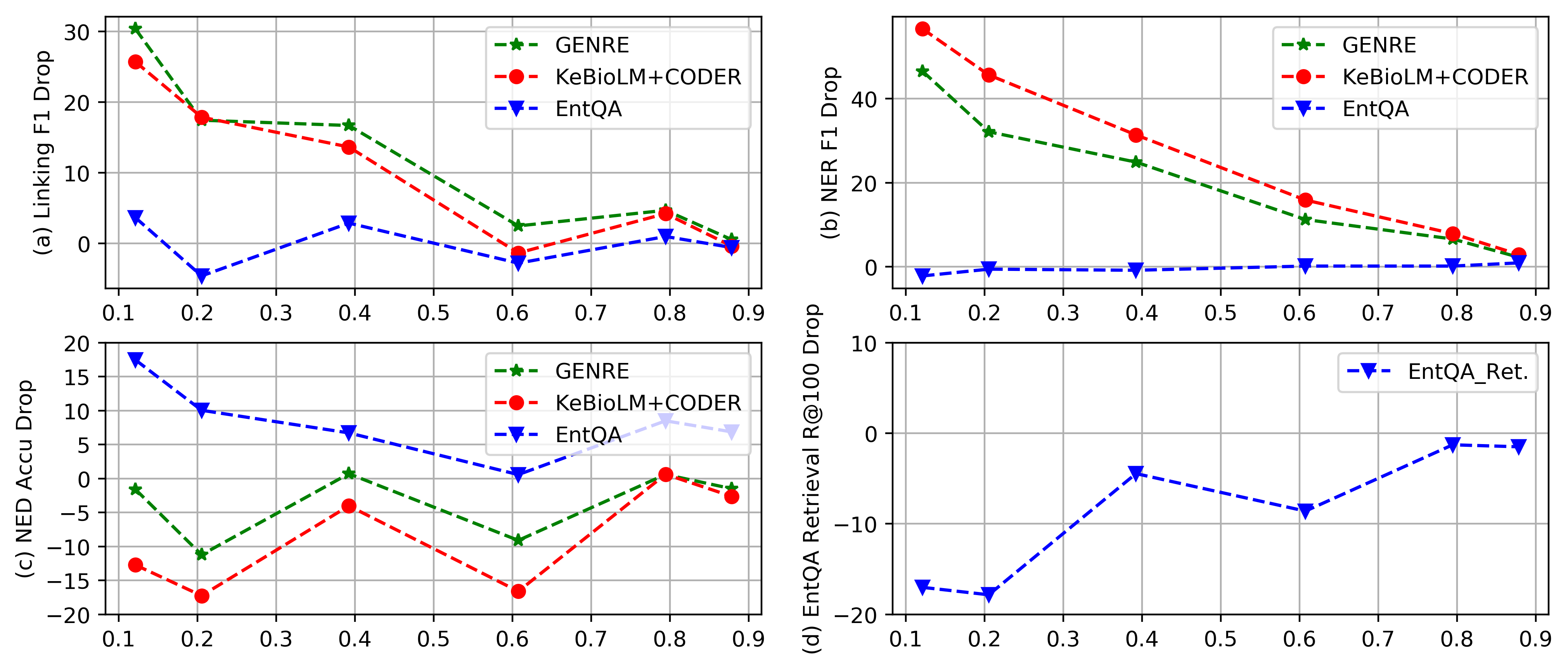}
    }
    \caption{The x-axis is the proportion of mention-concept annotations corresponding to the partial KBs in training data. The six points in each line represent different partial KBs in MedMentions. }
    \label{fig:f1drop}
    \vspace{-1em}
\end{figure*}

\paragraph{Redemptions Performances} \Cref{tab:main_transfer_enhance} shows results of partial KB inference on \MEDICint and \MEDICext on BC5CDR. 
We also identify the same pattern in other subset KBs (\Cref{app:redeem_result}).
Paradigms behave differently under these settings.

For KeBioLM+CODER, the best improvements are brought by thresholding.
Mention-concept pairs with low similarities can be categorized into concepts within $\mathcal{E}_1-\mathcal{E}_2$ or incorrect mention spans.
These two kinds of pairs are removed by thresholding which results of the improvement of NER and NED.
Post-pruning also brings improvement of NED by removing concepts within $\mathcal{E}_1-\mathcal{E}_2$, but it cannot deal with incorrect mention spans.

For EntQA, direct partial KB transfer achieves similar results to In-KB training.
The great performance of direct partial KB transfer is due to the integration of the thresholding mechanism.

For GENRE, the best performance is achieved uniformly by post-pruning. 
Post-pruning removes concepts within $\mathcal{E}_1-\mathcal{E}_2$ to boost performance.
Thresholding also has significant improvement and performs better than In-KB training.
The reason thresholding performs worse than post-pruning may be the log-likelihood is not a direct estimate of mention-entity pair validity.

Another observation is two redemption methods can outperform direct In-KB training, which suggests additional supervision from KB $\mathcal{E}_1-\mathcal{E}_2$ can benefit partial KB inference on $\mathcal{E}_2$.

\subsection{Discussion}
\label{sec:analysis}

In this section, we provide a further investigation into what causes performance variance across different partial KBs.
In training data, annotations associated with different partial KBs may take different proportions of total annotations.
Models may over-fit the frequency of mention annotations existing in training samples. 
We visualize F1 performance drop of entity linking and mention detection against the proportion of partial KB annotations in training data. 
As shown in \Cref{fig:f1drop}(a)(b), the performance drop is negatively correlated with the annotation proportions for GENRE and KeBioLM+CODER.
The relation is more prominent for mention detection.  
For EntQA, performances barely change in terms of entity linking and mention detection due to its robustness. 
This negative correlation suggests mention detection of GENRE and KeBioLM+CODER over-fit annotation frequency.
EntQA detects mentions according to retrieved concepts.
This explicit modeling makes it more robust since it handles out-of-KB mentions by filtering out irrelevant concepts in the retrieving stage.

For NED as shown in \Cref{fig:f1drop}(c), there exists no obvious trend between accuracy drops and annotation proportions.
For GENRE and KeBioLM+CODER, the disambiguation performances are improved when inference on partial KBs.
Improvements are also observed for EntQA on concept retrieval R@100. 
Concept spaces are shrunk for partial KBs and therefore the disambiguation problem becomes easier to approach. 
Contrarily, the disambiguation accuracy of EntQA drops, which is probably because of the distribution shift of retrieved concepts between training and inference
which serve as inputs for the reader. The distribution shifts in a way that for the same number of top retrieved concepts many concepts with lower ranks may be unseen for the reader in partial KB inference.
This illustrates EntQA is still influenced by partial KB inference although it is robust for detecting mentions.

\section{Conclusion}\label{sec:conclusion}

In this work, we propose a practical scenario, namely partial KB inference, in biomedical EL and give a detailed definition and evaluation procedures for it.
We review and categorize current state-of-the-art entity linking models into three paradigms.
Through experiments, we show NER-NED and simultaneous generation paradigms have vulnerable performance toward partial KB inference which is mainly caused by mention detection precision drop.
The NED-NER paradigm is more robust due to well-modeled mention-concept reliance.
We also propose two methods to redeem the performance drop in partial KB inference and discover out-KB annotations may enhance the in-KB performance.
Post-pruning and thresholding can both improve the performance of NER-NED and simultaneous generation paradigms. Although post-pruning is easy-to-use, it needs to store the large KB $\mathcal{E}_1$ (with their embeddings or trie) which has large memory consumption.
Thresholding does not rely on large KB $\mathcal{E}_1$ which also has better performance on the NER-NED paradigm.
Our findings illustrate the importance of partial KB inference in EL which shed light on the future research direction.

\section*{Limitations}
We only investigate representative methods of three widely-used EL paradigms.
However, there are more EL methods and paradigms we may not cover, and we leave them as future works.
Furthermore, more auxiliary information in the biomedical domain can be introduced to address the NIL issue we identify in this work.
For example, a hierarchical structure exists for concepts in KBs in the biomedical domain.
Therefore, NIL may be solved by linking them to hypernym concepts in the partial KBs \cite{nilinker}.
We consider the hierarchical mapping between NILs and in-KB concepts as a potential solution for performance degradation in partial KB inference.

Users can obtain different entity-linking results based on their own KBs which have the potential risk of missing important clinical information from the texts.

\section*{Ethics Statement}
Datasets used for building partial KB inference do not contain any patient privacy information.

\section*{Acknowledgement}

We would like to express our appreciation and gratitude to Professor Sheng Yu from Center for Statistical Science, Tsinghua University and Professor Muhao Chen from University of Southern California who have provided computational resources for this research. Vive l'amiti\'e parmi les auteurs.

\bibliography{anthology,custom}
\bibliographystyle{acl_natbib}

\appendix
\label{sec:appendix}

\section{Hyper-parameters}\label{app:train_hyper}

We demonstrates the hyper-parameters we used in training three EL models on MedMEntions and BC5CDR in \Cref{tab:model_train_hyper}.
All other hyper-parameters in training and inference that are not mentioned in this table are the same from the public codes and scripts of GENRE\footnote{Github repository of GENRE: \url{https://github.com/facebookresearch/GENRE}}, EntQA\footnote{Github repository of EntQA: \url{https://github.com/WenzhengZhang/EntQA}}, KeBioLM\footnote{Github repository of KeBioLM: \url{https://github.com/GanjinZero/KeBioLM}}, and CODER\footnote{Github repository of CODER: \url{https://github.com/GanjinZero/CODER}}. Models are implemented on single NVIDIA V100 GPU with 32GB memory. 

\begin{table*}[h]
\small 
\centering
\begin{tabular}{lcccc}
\toprule
& KeBioLM & GENRE & EntQA-retriever & EntQA-reader \\ 
\multicolumn{5}{c}{\textbf{BC5CDR}}\\

Train Length & 20 Epochs & 8000 Steps& 20 Epochs & 20 Epochs \\ 
Learning Rate & $1\times 10^{-5}$ & $3\times 10^{-5}$ & $2\times 10^{-6}$& $1\times 10^{-5}$\\
Warmup & 570 & 600 & 20\% & 6\%\\
Batch Size &16 & 8 & 8 & 2\\
Adam $\beta$ & (0.9,0.999) & (0.9,0.999)& (0.9,0.999)& (0.9,0.999) \\
Adam $\epsilon$ & $1\times 10^{-8}$ & $1\times 10^{-8}$& $1\times 10^{-8}$& $1\times 10^{-8}$\\
Weight Decay & 0.0 & 0.01 & 0 & 0\\
Clip Norm & 1.0 & 0.1 & - & -\\
Label Smoothing & 0.0 & 0.1 & - & -\\

\multicolumn{5}{c}{\textbf{MedMentions}}\\

Train Length & 20 Epochs & 8000 Steps & 50 Epochs & 50 Epochs\\ 
Learning Rate & $1\times 10^{-5}$ & $3\times 10^{-5}$ & $5\times 10^{-6}$& $1\times 10^{-5}$\\
Warmup & 570 & 600 & 20\% & 6\% \\
Batch Size &16 & 8 & 8 & 2\\
Adam $\beta$ & (0.9,0.999) & (0.9,0.999) & (0.9,0.999)& (0.9,0.999)\\
Adam $\epsilon$ & $1\times 10^{-8}$ & $1\times 10^{-8}$& $1\times 10^{-8}$& $1\times 10^{-8}$\\
Weight Decay & 0.0 & 0.01 & 0 & 0\\
Clip Norm & 1.0 & 0.1 & - & -\\
Label Smoothing & 0.0 & 0.1 & - & -\\

\bottomrule
\end{tabular}
\caption{The training settings for investigated models on BC5CDR and MedMentions. We leave out CODER as CODER is not further fine-tuned on downstream samples. }
\label{tab:model_train_hyper}
\small 
\end{table*}

\section{Datasets Statistics}\label{app:data_stat}

\Cref{tab:stats} shows the detailed statistics of data we used for partial KB inference. We use MeSH and MEDIC in the BC5CDR corpus\footnote{BC5CDR: \url{https://biocreative.bioinformatics.udel.edu/tasks/biocreative-v/track-3-cdr/}}. 
The BC5CDR dataset has been identified as being free of known restrictions under copyright law.
We use UMLS, MeSH and SNOMED from the 2017 AA release of UMLS.
To meet the assumption that MEDIC forms a subset of MeSH, we ditch the concepts in MEDIC that do not exist in MeSH.
And we use st21pv version of MedMentions\footnote{Github repository of MedMentions: \url{https://github.com/chanzuckerberg/MedMentions}}.
The MedMentions dataset is under CC0 licence.
We follow GenBioEL\footnote{Github repository of GenBioEL: \url{https://github.com/Yuanhy1997/GenBioEL}} for preprocessing the concepts and synonyms in the original KBs.
To meet the assumption that the partial KBs do not contain concepts out of training KB, we ditch the concepts in partial KBs that do not exist in UMLS. 

We use precision, recall, and F1 as metrics for entity linking and mention detection, and accuracy on correctly detected mentions for disambiguation performance. 
We also use the top 100 recall (R@100) to illustrate the performance of EntQA retriever.

\begin{table*}[h]
\small 
\centering
\setlength{\tabcolsep}{1.5mm}{
\begin{tabular}{lccccc}
\toprule
Target KB & \#Concepts & \#Annotations & \#Annotated Concepts & \#Annot. in Train & \#Concepts in Train \\
\midrule
\multicolumn{6}{c}{\textbf{BC5CDR}}\\
MeSH &268,146 & 9,269/9,511/9,655 & 1,304/1,246/1,299 & -/7,439/7,504 & -/681/691 \\
\MEDICint & 11,209 & 4,149/4,217/4,307 & 652/595/636 & -/3,526/3,655 & -/360/390 \\
\MEDICext & 256,937 & 5,120/5,294/5,348  & 652/651/663 & -/3,913/3,849 & -/321/301  \\
\midrule
\multicolumn{6}{c}{\textbf{MedMentions}} \\
UMLS & 2,368,641 & 122,241/40,884/40,157 & 18,520/8,643/8,457 & -/9,320/9,072& -/3,659/3,590 \\
\SNOMEDint & 342,998 & 74,272/25,385/24,391 & 9,678/4,768/4 ,716 & -/4,556/4,766 & -/1,807/1,779\\
\SNOMEDext &2,025,643 &47,969/15,499/15,766&8,842/3,875/3,741&-/4,554/4,516 & -/1,852/1,811 \\
\TAint & 184,939 & 25,109/8,240/8,117 & 3,805/1,797/1,741 & -/1,822/1,741 & -/734/721 \\
\TAext & 2,183,702 &97,132/32,644/32,040  &14,715/6,846/6,716  &  -/7,498/7,330 & -/2,925/2,869 \\
\TBint & 122,433 & 14,835/4,682/4,789 & 2,382/1,104/1,100 & -/1,000/1,051 & -/454/440 \\
\TBext & 2,246,208 &107,406/36,202/35,368  &16,138/7,539/7,357  &-/8,320/8,021  & -/3,205/3,150 \\
\bottomrule
\end{tabular}}
\caption{Dataset and corresponding knowledge base statistics.}
\label{tab:stats}
\small 
\end{table*}

\section{Appendix for Redemption Methods}

\subsection{Illustrative Example}\label{app:example}
We show an entity linking result on an example from BC5CDR:

\textit{Indomethacin induced {hypotension} in {sodium and volume} depleted rats. After a single oral dose of 4 mg/kg {indomethacin} ({IDM}) to {sodium} and volume depleted rats plasma renin activity (PRA) and systolic blood pressure fell significantly within four hours.}

The entity linking results are shown in Table \ref{tab:case}. In Post-Pruning, the final results (marked blue) are those linked to a concept in the partial KB MEDIC. In Thresholding, the final results (marked blue) are those scores larger than a fix threshold, for KeBioLM+CODER is 0.8, GENRE is -0.15 and EntQA is 0.043.

\begin{table*}[h]
    \centering
    \scriptsize
    \resizebox{1.0\textwidth}{!}{
    \begin{tabular}{l|ccc|cccc}
                \hline
                 \multirow{2}{*}{Methods}&\multicolumn{3}{c|}{Post-Pruning}&\multicolumn{4}{c}{Thresholding} \\
         & Mention Span & Concept & In Partial KB & Mention Span & Concept & Score & $\ge$Threshold \\
         \hline
        \multirow{6}{*}{Ke.+CO.}&(0,12)& D007213:indomethacin& False&(0,12)& C564365:ilvasc& 0.35& False \\
        &(21,32)& D007022:hypotension& \textcolor{blue}{True}&(21,32)& D007022:hypotension& 1.00& \textcolor{blue}{True} \\
        &(36,53)& D005441:fluids and secretions& False&(36,53)& D003681:water stress& 0.43& False \\
        &(105,117)& D007213:indomethacin& False&(105,117)& C564365:ilvasc& 0.35& False \\
        &(119,122)& D003922:iddm& \textcolor{blue}{True}&(119,122)& D003922:iddm& 0.94& \textcolor{blue}{True} \\
        &(127,133)& D012964:sodium& False&(127,133)& D000747:chloroses& 0.38& False \\
         \hline
         \multirow{5}{*}{GENRE}&(0,12)& D007213:amuno& False&(21,32)& D007022:hypotension& -0.067& \textcolor{blue}{True}\\
         &(21,32)& D007022:hypotension& \textcolor{blue}{True}&(36,42)& D007022:hypotension& -1.349& False \\
         &(36,42)& D012964:sodium& False &(105,117)& C563086:amc syndrome& -1.848& False \\
         &(105,117)& D007213:amuno& False &(127,133)& D007022:hypotension& -0.491& False  \\
         &(127,133)& D012964:sodium& False \\
         \hline
    \multirow{3}{*}{EntQA}&(0,12)& D007213:indomethacin& False&(0,12)& C564365:ilvasc& 0.087& \textcolor{blue}{True}  \\
        &(21,32)& D007022:hypotension& \textcolor{blue}{True}&(21,32)& D007022:hypotension& 0.098& \textcolor{blue}{True} \\
        &(127,133)& D012964:sodium& False&(127,133)& D000747:chloroses& 0.004& False \\
         \hline
    \end{tabular}
    }
    \caption{An illustrative example of Post-Pruning and Thresholding.}
    \label{tab:case}
\end{table*}

\subsection{Results on Different Datasets}\label{app:redeem_result}

\Cref{tab:main_transfer_enhance_app} shows results of the same experiments described in \Cref{sec:post-processing} on MedMentions with partial KB \SNOMEDint and \SNOMEDext.
Results on these table also supports the conclusion we provides at \Cref{sec:conclusion}.
We find thresholding and post-pruning benefit EntQA in this additional results whereas we witness a significantly performance drop in \Cref{tab:main_transfer_enhance}.
This suggests performance of thresholding and post-pruning on EntQA is different across partial KBs.
Nevertheless, we have not seen a dramatic performance boost (as those in GENRE and KeBioLM+CODER) brought by post-processing techniques on EntQA.

\begin{table*}[h]
\small 
\centering
\setlength{\tabcolsep}{1.2mm}{
\begin{tabular}{p{2mm}l|cccc|cccc}
\toprule
&& \multicolumn{4}{c}{\SNOMEDint}\vline & \multicolumn{4}{c}{\SNOMEDext} \\
&& EL-P/R&EL-F1&NER-F1&NED-Acc& EL-P/R&EL-F1&NER-F1&NED-Acc \\
\hline
\cellcolor{blue!10}& In-KB Train &53.10/26.28&\underline{35.16}&64.37&76.92&44.77/23.99&\textbf{31.24}&\textbf{67.40} & \textbf{70.12}\\
\cellcolor{blue!10}& Partial KB Inference &46.04/27.01&34.05&63.54&74.81& 36.75/23.12&28.38 & 64.53 & 68.67 \\
\cellcolor{blue!10}& \ \ +w/ Thresholding &44.92/30.01&\textbf{35.98} &\textbf{64.98}&\textbf{79.10}& 35.10/27.33&\underline{30.73}&\underline{65.73} & \underline{69.04} \\
\cellcolor{blue!10}\multirow{-4}{*}{\rotatebox[origin=c]{90}{EntQA}}& \ \ +w/ Post-pruning&44.70/28.23&34.45&\underline{64.77}&\underline{78.82}&35.48/26.92&30.61 &65.71&68.88\\
\hline
\cellcolor{blue!10}& In-KB Train &46.15/39.95&42.83&53.38&\textbf{80.21}&42.24/25.68&31.94 &44.77&\underline{71.35}\\
\cellcolor{blue!10}& Partial KB Inference &34.40/49.40&40.56&53.97 &75.14& 19.82/39.28&26.35&40.33 &65.33 \\
\cellcolor{blue!10}& \ \ +w/ Thresholding & 45.25/44.25&\underline{44.75}&\underline{56.38} &\underline{79.37}&30.45/34.09&\underline{32.17}&\underline{45.13}&71.26\\
\cellcolor{blue!10}\multirow{-4}{*}{\rotatebox[origin=c]{90}{GENRE}}& \ \ +w/ Post-pruning & 46.20/47.18&\textbf{46.68}&\textbf{59.05}&79.07& 38.27/36.56&\textbf{37.39}&\textbf{51.64}&\textbf{72.41}\\
\hline
\cellcolor{blue!10}& In-KB Train  &43.87/44.62&\underline{44.24}&\textbf{64.00}&69.13&30.62/30.36&\textbf{30.49}&\textbf{55.41}&55.02 \\
\cellcolor{blue!10}& Partial KB Inference &28.19/48.28&35.59&54.58&65.22&14.18/37.54&20.59 &39.11 &52.64\\
\cellcolor{blue!10}& \ \ +w/ Thresholding &53.48/44.64&\textbf{48.66}&\underline{61.42}&\textbf{79.23}&28.54/31.33&\underline{29.87}&44.37&\textbf{67.33}  \\
\cellcolor{blue!10}\multirow{-4}{*}{\rotatebox[origin=c]{90}{Ke.+CO.}}& \ \ +w/ Post-pruning&46.86/33.73&39.22&51.44&\underline{76.26}&23.65/36.27&28.63&\underline{50.06}&\underline{57.19}\\
\bottomrule
\end{tabular}
}
\caption{Additional Results of partial KB inference, In-KB training and two redemption methods for three investigated models. The results are evaluated on partial KB \SNOMEDint and \SNOMEDext in MedMentions.
The best performance for a model in each dataset is identified with \textbf{bold} and the second is \underline{underlined}.}
\label{tab:main_transfer_enhance_app}
\small 
\end{table*}

\end{document}